\documentclass{article}
\usepackage{style/foci}
\usepackage{tgpagella}   
\usepackage{tgheros}     
\usepackage{flafter}
\usepackage{comment}
\usepackage{array}
\usepackage{tikz}
\usetikzlibrary{arrows.meta,positioning,fit}
\newfloat{algorithm}{tbp}{loa}
\floatname{algorithm}{Algorithm}

\usepackage[utf8]{inputenc}
\usepackage[T1]{fontenc}
\usepackage{microtype}

\usepackage{amsmath}
\usepackage{amssymb}
\usepackage{amsfonts}
\usepackage{amsthm}
\usepackage{mathtools}
\usepackage{mathrsfs}
\usepackage{physics}
\usepackage{braket}
\usepackage{slashed}
\usepackage{nicefrac}
\usepackage{textcomp}
\usepackage{dsfont}
\usepackage{bbm}
\usepackage{bm}

\usepackage{graphicx}
\usepackage{subcaption}
\usepackage[export]{adjustbox}
\usepackage{float}
\usepackage{booktabs}
\usepackage{dcolumn}
\newcolumntype{d}[1]{D{.}{.}{#1}}
\usepackage{bigstrut, tabularx, multirow, makecell, diagbox}
\usepackage{colortbl}
\usepackage{tabularray}
\UseTblrLibrary{booktabs}
\usepackage{threeparttable}
\usepackage{tablefootnote}
\usepackage{fontawesome5}

\usepackage{placeins}
\usepackage{caption}
\usepackage{footnote}
\usepackage{enumitem}
\usepackage{multicol}
\usepackage{xspace}
\usepackage{titletoc}
\usepackage{titlesec}
\usepackage[bottom]{footmisc}
\usepackage{setspace}

\usepackage{wrapfig}
\usepackage{tikz}
\usepackage{quantikz}
\usepackage{dashbox}
\usepackage{mdframed}
\usepackage{marvosym}
\usepackage{pifont}
\usepackage{CJK}
\usepackage{url}

\usepackage[table,x11names]{xcolor}
\usepackage[most]{tcolorbox}
\tcbuselibrary{breakable}
\usetikzlibrary{decorations.pathreplacing, fit}

\definecolor{primaryblue}{HTML}{0066CC}
\definecolor{accentcyan}{HTML}{00D4AA}
\definecolor{warmorange}{HTML}{FF6B35}
\definecolor{deepgray}{HTML}{2C3E50}
\definecolor{lightgray}{HTML}{F8F9FA}
\definecolor{gradientstart}{HTML}{667eea}
\definecolor{gradientend}{HTML}{764ba2}

\definecolor{citecolor}{HTML}{0071bc}
\definecolor{citeblue}{RGB}{0, 113, 188}
\definecolor{linkcolor}{HTML}{9A4D92}
\definecolor{firebrick}{rgb}{0.698,0.133,0.133}

\definecolor{paleviolet}{HTML}{E1EEFC}
\definecolor{CarolinaUltraLight}{HTML}{E7F4FC}
\definecolor{lightgrey}{RGB}{247, 247, 247}
\definecolor{shadecolor}{HTML}{EFEFEF}
\definecolor{lightyellow}{rgb}{1.0, 0.95, 0.7}
\definecolor{lightblue}{rgb}{0.90, 0.95, 1.0}
\definecolor{light-gray}{gray}{0.95}

\definecolor{darkgrey}{rgb}{0.5, 0.5, 0.5}
\definecolor{darkgreen}{rgb}{0, 0.5, 0}
\definecolor{mydarkblue}{rgb}{0,0.08,0.45}
\definecolor{mydarkblue2}{rgb}{0.133, 0.133, 0.698}
\definecolor{echodrk}{HTML}{0099cc}
\definecolor{mymauve}{rgb}{0.58,0,0.82}
\definecolor{midnightblue}{rgb}{0.1,0.1,0.44}
\definecolor{oxfordblue}{rgb}{0.0,0.13,0.28}
\definecolor{prussianblue}{rgb}{0.0,0.19,0.33}
\definecolor{coolteal}{rgb}{0, 0.45, 0.45}
\definecolor{olive}{rgb}{0.1, 0.3, 0}
\definecolor{mypurple}{rgb}{0.5,0,0.5}
\definecolor{almond}{rgb}{0.94, 0.87, 0.8}

\definecolor{blue_ampEncoding}{HTML}{DAE8FC}
\definecolor{green_encoder}{HTML}{D5E8D4}
\definecolor{purple_decoder}{HTML}{E1D5E7}
\definecolor{yellow_measure}{HTML}{FFF2CC}
\definecolor{gray_block}{HTML}{F5F5F5}
\definecolor{pink_dru}{HTML}{FAD9D5}
\definecolor{orange_v}{HTML}{FAD7AC}

\definecolor{colorA}{rgb}{1,0,0}
\definecolor{colorB}{rgb}{0,0.3,1}
\definecolor{colorC}{rgb}{0.9,0.8,0.2}
\definecolor{colorD}{rgb}{0,0.65,0}
\definecolor{lesslightgray}{rgb}{0.5,0.5,0.5}
\definecolor{fundamental}{RGB}{55, 110, 111}
\definecolor{Gred}{RGB}{219, 50, 54}
\definecolor{ToCgreen}{RGB}{0, 128, 0}
\definecolor{Sepia}{RGB}{112, 66, 20}
\definecolor{Dblue}{rgb}{0,0.08,0.45}
\definecolor{Blue}{rgb}{0, 0, 0.8}
\definecolor{blue}{rgb}{0,0,1}
\definecolor{UNCblue!10}{rgb}{0.84,0.91,0.98}
\definecolor{RowAlt}{rgb}{0.98,0.98,0.99}

\definecolor{CarolinaBlue}{HTML}{7BAFD4}        
\definecolor{CarolinaLightBlue}{HTML}{B3D4E5}   
\definecolor{CarolinaUltraLight}{HTML}{E8F4F8}  
\definecolor{CarolinaText}{HTML}{1C2B33}        

\usepackage[pagebackref=true,breaklinks=true,colorlinks,hyperfootnotes=false]{hyperref}
\hypersetup{
  colorlinks,
  citecolor=citeblue,
  linkcolor=firebrick,
  urlcolor=firebrick
}
\usepackage[nameinlink,capitalize,noabbrev]{cleveref}

\titlespacing\section{0pt}{4pt plus 4pt minus 2pt}{-2pt plus 2pt minus 2pt}
\titlespacing\subsection{0pt}{2pt plus 4pt minus 2pt}{-2pt plus 2pt minus 2pt}
\titlespacing\subsubsection{0pt}{2pt plus 4pt minus 2pt}{-2pt plus 2pt minus 2pt}

\makeatletter
\def\th@remark{%
  \thm@headfont{\bfseries}%
  \normalfont 
  \thm@preskip\topsep \divide\thm@preskip\tw@
  \thm@postskip\thm@preskip
}
\makeatother

\theoremstyle{definition}

\tcbset{focithm/.style={
  enhanced, breakable, frame hidden,
  colback=black!5,
  width=\linewidth,
  enlarge left by=0pt, enlarge right by=0pt,
  boxsep=6pt, boxrule=0pt,
  left=4pt, right=4pt, top=1pt, bottom=1pt,
  arc=8pt,
  before skip=\topsep, after skip=\topsep
}}

\tcolorboxenvironment{theorem}{focithm}

\tcolorboxenvironment{lemma}{focithm}

\tcolorboxenvironment{corollary}{focithm}

\tcolorboxenvironment{proposition}{focithm}

\tcolorboxenvironment{definition}{focithm}

\tcolorboxenvironment{assumption}{focithm}

\tcolorboxenvironment{claim}{focithm}

\tcolorboxenvironment{problem}{focithm}

\tcolorboxenvironment{question}{focithm}

\newtcolorbox{titleblock}{
  enhanced,
  frame hidden,
  colback=CarolinaUltraLight,
  colframe=CarolinaUltraLight,
  boxrule=0pt,
  arc=10pt,
  left=14pt,
  right=14pt,
  top=14pt,
  bottom=14pt,
  width=\linewidth,
  before skip=12pt plus 4pt,
  after skip=12pt plus 4pt,
  grow to left by=1.5pt,
  grow to right by=1.5pt,
  before upper={
    \setlength{\parindent}{0cm}
    \setlength{\parskip}{0.5cm}
  }
}

\crefname{theorem}{Theorem}{Theorems}
\crefname{proposition}{Proposition}{Propositions}
\crefname{lemma}{Lemma}{Lemmas}
\crefname{corollary}{Corollary}{Corollaries}
\crefname{definition}{Definition}{Definitions}
\crefname{assumption}{Assumption}{Assumptions}
\crefname{remark}{Remark}{Remarks}
\crefname{problem}{Problem}{Problems}
\crefname{property}{Property}{property}
\crefname{question}{Question}{Questions}

\numberwithin{equation}{section}
\numberwithin{theorem}{section}
\numberwithin{proposition}{section}
\numberwithin{definition}{section}
\numberwithin{lemma}{section}
\numberwithin{assumption}{section}
\numberwithin{remark}{section}

\def\1{\bm{1}}

\makeatletter
\let\save@mathaccent\mathaccent
\newcommand*\if@single[3]{%
    \setbox0\hbox{${\mathaccent"0362{#1}}^H$}%
    \setbox2\hbox{${\mathaccent"0362{\kern0pt#1}}^H$}%
    \ifdim\ht0=\ht2 #3\else #2\fi
}
\newcommand*\rel@kern[1]{\kern#1\dimexpr\macc@kerna}
\newcommand*\widebar[1]{\@ifnextchar^{{\wide@bar{#1}{0}}}{\wide@bar{#1}{1}}}
\newcommand*\wide@bar[2]{\if@single{#1}{\wide@bar@{#1}{#2}{1}}{\wide@bar@{#1}{#2}{2}}}
\newcommand*\wide@bar@[3]{%
    \begingroup
    \def\mathaccent##1##2{%
        \let\mathaccent\save@mathaccent
        \if#32 \let\macc@nucleus\first@char \fi
        \setbox\z@\hbox{$\macc@style{\macc@nucleus}_{}$}%
        \setbox\tw@\hbox{$\macc@style{\macc@nucleus}{}_{}$}%
        \dimen@\wd\tw@
        \advance\dimen@-\wd\z@
        \divide\dimen@ 3
        \@tempdima\wd\tw@
        \advance\@tempdima-\scriptspace
        \divide\@tempdima 10
        \advance\dimen@-\@tempdima
        \ifdim\dimen@>\z@ \dimen@0pt\fi
        \rel@kern{0.6}\kern-\dimen@
        \if#31
        \overline{\rel@kern{-0.6}\kern\dimen@\macc@nucleus\rel@kern{0.4}\kern\dimen@}%
        \advance\dimen@0.4\dimexpr\macc@kerna
        \let\final@kern#2%
        \ifdim\dimen@<\z@ \let\final@kern1\fi
        \if\final@kern1 \kern-\dimen@\fi
        \else
        \overline{\rel@kern{-0.6}\kern\dimen@#1}%
        \fi
    }%
    \macc@depth\@ne
    \let\math@bgroup\@empty \let\math@egroup\macc@set@skewchar
    \mathsurround\z@ \frozen@everymath{\mathgroup\macc@group\relax}%
    \macc@set@skewchar\relax
    \let\mathaccentV\macc@nested@a
    \if#31
    \macc@nested@a\relax111{#1}%
    \else
    \def\gobble@till@marker##1\endmarker{}%
    \futurelet\first@char\gobble@till@marker#1\endmarker
    \ifcat\noexpand\first@char A\else
    \def\first@char{}%
    \fi
    \macc@nested@a\relax111{\first@char}%
    \fi
    \endgroup
    }
\makeatother

\DeclareMathAlphabet{\mathsfit}{\encodingdefault}{\sfdefault}{m}{sl}
\SetMathAlphabet{\mathsfit}{bold}{\encodingdefault}{\sfdefault}{bx}{n}

\newcommand{\R}{\mathbb{R}}
\newcommand{\C}{\mathbb{C}}

\newcommand{\E}{\mathbb{E}}

\let\hat\widehat

\makeatletter

\def\foci@authors{}
\def\foci@affil{}
\def\foci@email{}
\def\foci@project{}
\def\foci@date{\today}
\def\foci@authorsep{}

\newcommand{\fociauthor}[2]{%
  \g@addto@macro\foci@authors{\foci@authorsep\mbox{#1\textsuperscript{#2}}\gdef\foci@authorsep{,\ }}}
\newcommand{\fociaffil}[1]{\gdef\foci@affil{#1}}
\newcommand{\fociemail}[1]{\gdef\foci@email{#1}}
\newcommand{\fociproject}[1]{\gdef\foci@project{#1}}
\newcommand{\focidate}[1]{\gdef\foci@date{#1}}
\newcommand{\fociauthorlist}{\gdef\foci@authorsep{}\foci@authors}

\def\foci@blfootnote{\gdef\@thefnmark{}\@footnotetext}
\newcommand{\focifootnotes}{%
  \foci@blfootnote{\textsuperscript{*}Equal contribution.\quad
    \textsuperscript{\Letter}Corresponding author: \texttt{\foci@email}}}

\newlength{\foci@mw}
\newcommand{\focidrawmark}[3]{%
  \setlength{\foci@mw}{#2}%
  \begin{scope}[shift={#3}, x={\foci@mw/564.03}, y={-\foci@mw/564.03}]
    \foreach \cx/\cy in {282.02/66.03, 282.02/222.08, 215.77/348.8, 348.73/348.8,
                         414.88/474.93, 149.16/474.93, 67.06/222.08, 496.99/222.08}
      \fill[#1] (\cx,\cy) circle[radius=42.04];
  \end{scope}}
\newcommand{\focimark}[2][black]{%
  \setlength{\foci@mw}{#2}%
  \begin{tikzpicture}[x={\foci@mw/564.03}, y={-\foci@mw/564.03}]
    \useasboundingbox (0,0) rectangle (564.03,540.96);
    \focidrawmark{#1}{#2}{(0,0)}
  \end{tikzpicture}}

\fancypagestyle{focititlepage}{\fancyhf{}\renewcommand{\headrulewidth}{0pt}}

\newcommand{\foci@headerline}{%
  \noindent
  \begin{minipage}[c]{0.5\linewidth}
    \includegraphics[height=0.58cm]{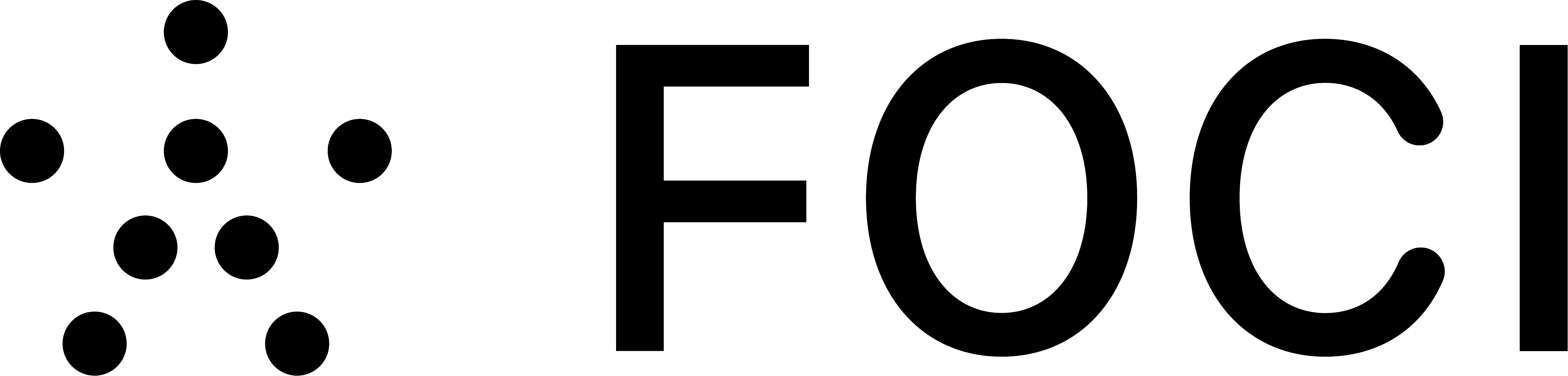}
  \end{minipage}%
  \begin{minipage}[c]{0.5\linewidth}
    \raggedleft\small\itshape\color{black!55}\foci@date
  \end{minipage}\par}

\newcommand{\foci@contactline}{%
  {\footnotesize\color{black!65}%
   \textbf{Email:}~\texttt{\foci@email}\qquad
   \textbf{Project:}~\url{\foci@project}\par}}

\newcommand{\foci@hairline}{\noindent\textcolor{black}{\rule{\linewidth}{0.5pt}}\par}

\newcommand{\focicoverCard}{%
  \thispagestyle{focititlepage}%
  \foci@headerline
  \vskip -0.5em
  \foci@hairline
  \vskip 0.9em
  \begin{tcolorbox}[enhanced, frame hidden, boxrule=0pt, arc=12pt,
      colback=black!4, width=\linewidth,
      left=18pt, right=18pt, top=18pt, bottom=16pt,
      grow to left by=2pt, grow to right by=2pt]
    \noindent
    \begin{minipage}[t]{\linewidth}
      {\setstretch{1.15}\LARGE\bfseries\raggedright AutoAdapt: Reliable Few-Shot Adaptation under Clinical Distribution Shifts
\unskip\par}
      \vskip 0.55em
      {\normalsize\fociauthorlist\par}
      \vskip 0.2em
      {\small\color{black!70}\foci@affil\par}
    \end{minipage}\par
    \vskip 0.9em
    {\small\setstretch{1.08}Large pretrained clinical models provide a practical way to reuse learned prior knowledge across hospitals by adapting models to them.
In practice, a target hospital may only have a small labeled patient cohort, a setting commonly referred to as few-shot adaptation.
This requires making multiple decisions, such as which pretrained model to adapt, how much of the model to update, and which patients to use.
Nevertheless, this process faces two primary challenges. First, the best adaptation strategy varies across clinical tasks. Second, evaluating and comparing candidate strategies becomes unreliable due to the small patient cohort.
In this work, we introduce \method{} with two core designs to deal with these challenges.
The \emph{Adapter} defines an extensible space of adaptation recipes, and the \emph{Automator} forms a weighted recipe combination from evidence within the adaptation patients.
We propose a reliability rule to ensure that only the most effective strategy on most available patients will be selected. These selected strategies then form a combination for effective few-shot adaptation.
We conduct extensive experiments across critical care, emergency care, and diagnostic datasets, and the results show that \method{} consistently achieves state-of-the-art performance using only a few patients for adaptation.
\unskip\par}
    \vskip 0.9em
    \foci@contactline
  \end{tcolorbox}
  \vskip 0.4em}

\newcommand{\focicover}{\focicoverCard}

\makeatother

\newcommand{\method}{\textsc{AutoAdapt}}

\def\R{\mathcal{R}}
\def\C{\mathcal{C}}
\def\A{\mathcal{A}}
\def\E{\mathcal{E}}

\newcolumntype{L}[1]{>{\raggedright\arraybackslash}p{#1}}
\newcolumntype{C}[1]{>{\centering\arraybackslash}p{#1}}
\newif\ifselectedfiveanalysisavailable
\selectedfiveanalysisavailabletrue

\newif\ifselecteduniquepatientavailable
\selecteduniquepatientavailabletrue

\newif\ifselectedcrossexecutionavailable
\selectedcrossexecutionavailabletrue

\newif\ifselecteddabaselinesavailable
\selecteddabaselinesavailabletrue

\newif\ifselectedcheckpointgridavailable
\selectedcheckpointgridavailabletrue

\newif\ifselectedwindowweightingavailable
\selectedwindowweightingavailabletrue

\newif\ifselectedsignalavailabilityavailable
\selectedsignalavailabilityavailabletrue

\newif\ifselectedehrdocumentationavailable
\selectedehrdocumentationavailabletrue

\newif\ifselectedehrrecencyavailable
\selectedehrrecencyavailabletrue

\newif\ifselectedmodalitydisagreementavailable
\selectedmodalitydisagreementavailabletrue

\newif\ifselectedmodalitydirectionavailable
\selectedmodalitydirectionavailabletrue

\newif\ifselectedclinicalfusionavailable
\selectedclinicalfusionavailabletrue

\newif\ifselectedsimplerulereliabilityavailable
\selectedsimplerulereliabilityavailabletrue

\newif\ifselectedcarehistoryavailable
\selectedcarehistoryavailabletrue

\newif\ifselectedorgansupportavailable
\selectedorgansupportavailabletrue

\newif\ifselectedpreindexinstabilityavailable
\selectedpreindexinstabilityavailabletrue

\newif\ifselectedeventtimingavailable
\selectedeventtimingavailabletrue

\newif\ifselectedmonitoringworkloadavailable
\selectedmonitoringworkloadavailabletrue

\newif\ifselectedshiftavailable
\selectedshiftavailabletrue

\newif\ifselectedoperatingpointavailable
\selectedoperatingpointavailabletrue

\newif\ifselectedcalibrationavailable
\selectedcalibrationavailabletrue

\newif\ifselectedalertbudgetavailable
\selectedalertbudgetavailabletrue

\newif\ifselectedselectivetriageavailable
\selectedselectivetriageavailabletrue

\newif\ifselectedreviewworkloadavailable
\selectedreviewworkloadavailabletrue

\newif\ifselecteddecisioncurveavailable
\selecteddecisioncurveavailabletrue

\newif\ifselectedclinicalprofileavailable
\selectedclinicalprofileavailabletrue

\newif\ifselecteddemographicavailable
\selecteddemographicavailabletrue

\newif\ifselectedadmissionworkflowavailable
\selectedadmissionworkflowavailabletrue

\newif\ifselectedcomorbidityavailable
\selectedcomorbidityavailabletrue

\newif\ifselectedmultisystemavailable
\selectedmultisystemavailabletrue

\begin{document}

\fociauthor{Song Wang}{}
\fociauthor{Jie Peng}{}
\fociauthor{Davis Hobley}{}
\fociauthor{Zachary Plotkin}{}
\fociauthor{Tianlong Chen}{}

\fociaffil{Foci}
\fociemail{song@fo.ci, jie@fo.ci, davis@fo.ci, zach@fo.ci, tianlong@fo.ci}
\fociproject{https://fo.ci/}

\pagestyle{fancy}
\fancyhf{}
\renewcommand{\headrulewidth}{0.4pt}
\chead{\small\scshape AutoAdapt: Few-Shot Clinical Adaptation
\unskip}
\cfoot{\small\thepage}

\focicover

\section{Introduction}

\begin{wrapfigure}{r}{0.44\textwidth}
  \centering
  \vspace{-1.8em}
  \includegraphics[width=0.98\linewidth]{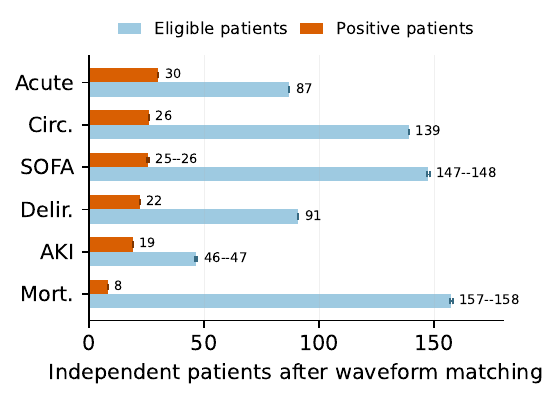}
    \vspace{-1.4em}
  \caption{Independent patient support in the  MIMIC-IV cohorts.}
  \label{fig:scarcity-teaser}
    \vspace{-.8em}
\end{wrapfigure}

Clinical models often face distribution shifts when deployed to new hospitals, where differences in patient data can degrade model performance. Although adapting models trained on other datasets to the target hospital can mitigate such shifts, labeled data from the new target distribution are often scarce. Particularly, clinical outcomes can be costly to annotate or unavailable for many patients, leaving only a small labeled cohort for adaptation. This limitation persists even in large clinical datasets, as many outcomes (e.g., acute kidney injury or vasopressor initiation) provide only one independent label per patient~\citep{johnson2016mimic3,johnson2023mimic4,moody2022mimic4waveform}. As an example, in the MIMIC-IV cohorts in Figure~\ref{fig:scarcity-teaser},  only 8--30 positive patients are available per task.

Pretrained physiological models provide a possible way for adaptation by transferring knowledge learned from large source datasets~\citep{kiyasseh2021clocs,gopal2021threeKG}. However, how to adapt a pretrained model to a new target distribution remains an important and complex decision. 
For example, it requires choosing which source checkpoint to transfer and how much of the model to update, which ranges from using a fixed encoder to partial or full parameter fine-tuning. 
The effectiveness of these choices can vary substantially across target tasks and distributions. Particularly, a specialized checkpoint may transfer well to a related outcome but poorly to another, while aggressive updating may help when sufficient labeled data is available but overfit when labels are scarce. Thus, there is no universally effective adaptation strategy for a new clinical target.


This creates two challenges. The first is \emph{adaptation heterogeneity}: the appropriate checkpoint and update rule can vary with the target clinical task and distribution. As a result, an adaptation strategy that works well for one target may perform poorly on another, making a fixed adaptation choice unreliable across different distributions.
The second is \emph{selection uncertainty}: determining which strategy to use requires comparing candidates using a small cohort  for adaptation. With only a few labeled patients, these comparisons can be highly unstable, and the apparent advantage of one strategy may be driven by only a small number of patients. 
Together, these challenges make reliable adaptation difficult when target data is limited~\citep{bendavid2010domain,zhang2021clinicaldg,guo2022temporalshift}.

In this work, we introduce \method{} to address these challenges. It consists of two components: (1) The \emph{Adapter} addresses adaptation heterogeneity by jointly composing choices of checkpoint, update, data, and input into complete recipes, capturing their interactions.
(2)  The \emph{Automator} addresses selection uncertainty by comparing recipes based on the limited target data, while removing unreliable choices when the evidence is insufficient.  Together, \method{} enables task-specific adaptation under scarce labels.
Our main contributions are summarized as follows:
\begin{itemize}[leftmargin=*]
\item We introduce an extensible \emph{Adapter} that represents adaptation strategies as complete recipes, jointly specifying the source checkpoint, update rule, target data policy, and input route. This allows different adaptation choices to be considered together rather than optimized independently.
\item We develop a reliability-aware \emph{Automator} that uses held-out patient predictions, task information, and label-free shift measures to construct a task-specific weighted recipe combination.
\item We evaluated \method{} across the ICU transfer, emergency care, and diagnostic ECG datasets. The results demonstrate improvements over fixed and existing adaptation strategies, including in settings with only one to six positive adaptation patients.
\end{itemize}

\section{Related Work}

Clinical time-series benchmarks made ICU prediction reproducible by fixing
cohorts, observation windows, and outcomes~\citep{harutyunyan2019multitask}.
Physiological pretraining asks a complementary question: whether structure
learned without target labels can support new tasks.  CLOCS and 3KG exploit
cross-view or temporal agreement in biosignals
~\citep{kiyasseh2021clocs,gopal2021threeKG}, while general time-series models
seek one representation that transfers broadly~\citep{goswami2024moment}.
These methods make source knowledge available.  They do not determine which
source checkpoint should be trusted for a particular target outcome.

Domain adaptation studies distinguish a change in observed inputs from a
change in the relation between inputs and labels~\citep{bendavid2010domain}.
This distinction is especially important in medicine because performance can
vary with site, time, and care process
~\citep{zhang2021clinicaldg,guo2022temporalshift}.  Frozen transfer, partial
fine-tuning, LoRA, and full fine-tuning make different assumptions about how
much of that relation remains stable~\citep{hu2022lora}.  They are candidate
responses to shift rather than a rule for deciding which response is supported
by a small target cohort.

Several lines of work address parts of this decision.  PRAM retrieves related
local patients~\citep{jeong2026pram}, and OTTEHR transports structured EHR
features across domains~\citep{li2026ottehr}.  LEEP, LogME, and TransRate score
the transferability of a representation
~\citep{nguyen2020leep,you2021logme,huang2022transrate}.  AutoPEFT searches over
parameter-efficient updates~\citep{zhou2024autopeft}, while TEMPLATE ranks
pretrained time-series models~\citep{zhang2025template}.  These approaches
typically optimize one component of transfer or presume that the observed
ranking is reliable.  \method{} instead compares complete recipes and treats
the reliability of that comparison as part of the adaptation problem.  It can
use a specialist, update the model, or add a modality when the patient-level
evidence supports that choice, and otherwise retain a predefined default
recipe.

\section{Problem Setting}
\label{sec:setting}

We consider a target clinical prediction task $t$ at a new site.  Let $V$
denote physiological waveforms, $Z$ denote optional structured EHR variables,
and $Y_t$ denote the target outcome.  The available source assets are a finite
library $\C=\{c_1,\ldots,c_K\}$ of pretrained checkpoints.  A checkpoint may
be a general waveform encoder or an encoder later specialized for a source
outcome.  Checkpoint $c_k$ was learned under source distribution
$P_{s_k}(V,Z,Y_k)$, whereas the target follows $P_t(V,Z,Y_t)$.  Transfer may
therefore involve both input shift and a change in the clinical prediction
target:
\begin{equation}
  P_{s_k}(V,Z)\neq P_t(V,Z),
  \qquad
  P_{s_k}(Y_k\mid V,Z)\neq P_t(Y_t\mid V,Z).
  \label{eq:shift}
\end{equation}
The first inequality captures changes in population, sensors, and acquisition.
The second captures the fact that a source outcome may only be partially
related to $Y_t$.  AutoAdapt measures these two forms of compatibility
separately.

Each target task specifies an index time $\tau$, observation length $O_t$, gap
$G_t$, and prediction horizon $H_t$.  For patient $i$, the input contains
waveform windows $X_i=(x_{ij})_{j=1}^{m_i}$ and optional structured context
$z_i$ observed in $[\tau-O_t,\tau]$.  The binary label $y_i$ records whether
the target event occurs in $(\tau+G_t,\tau+G_t+H_t]$.  The model outputs one
patient risk $\hat p_i=\hat f_t(X_i,z_i)$ after aggregating repeated-window
scores.  Task-specific values of $O_t$, $G_t$, and $H_t$ are reported in
Appendix Table~\ref{tab:cohort}.

The target patients are divided into a small labeled adaptation cohort $\A_t$
and a patient-disjoint evaluation cohort $\E_t$.  All patient-specific model
fitting and numerical combination-weight estimation use $\A_t$.  Higher-level
structural choices are treated separately, and their selection scope is stated
in the experimental protocol.  Because windows from one patient are
correlated, the patient is the evidence unit for fitting, internal validation,
and uncertainty.  Evaluation labels do not fit candidate predictors or
numerical combination weights.

\begin{figure*}[!t]
\centering
\includegraphics[page=1,width=.98\textwidth]{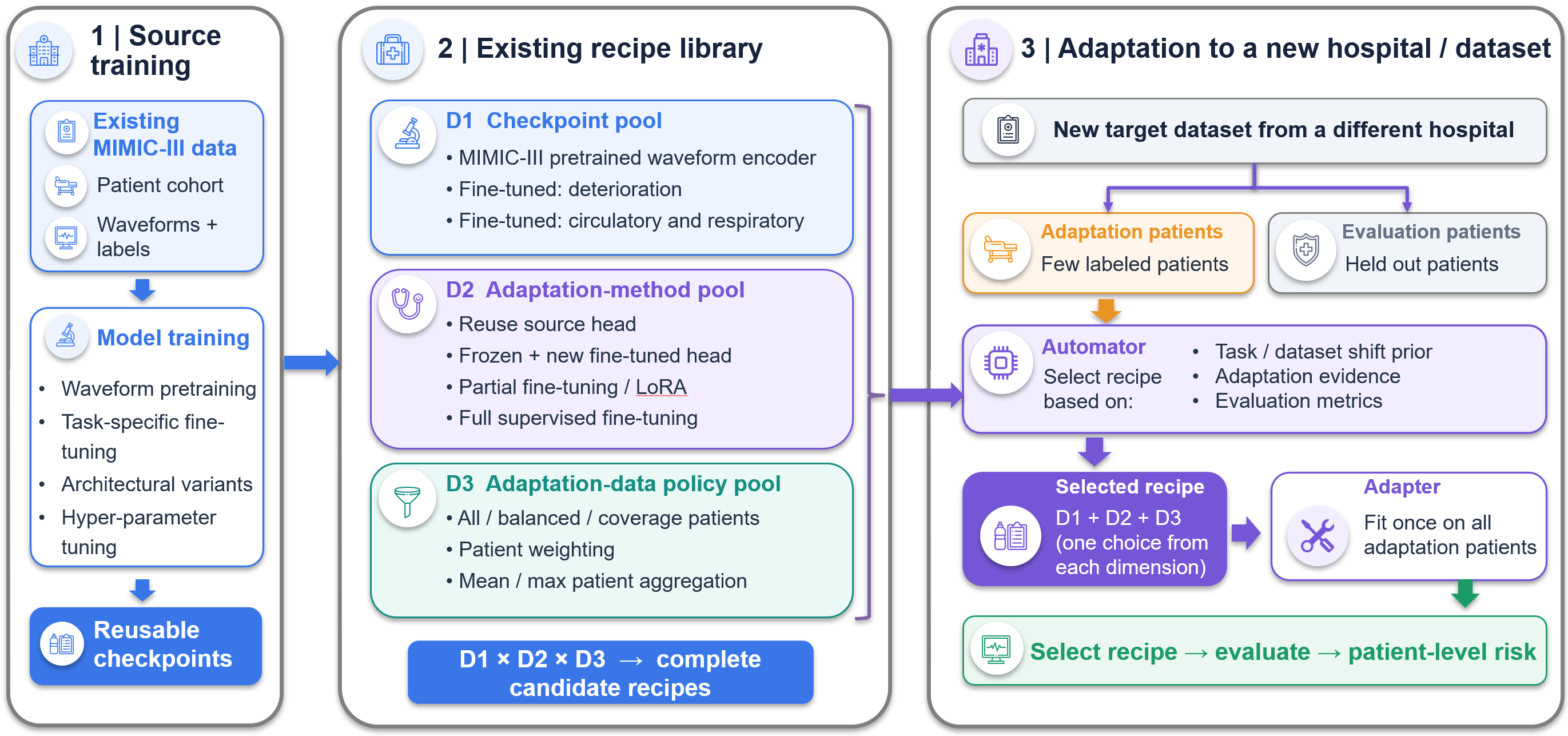}
\caption{Overview of \method{}.  The Adapter defines a recipe through its
source representation, update capacity, target data policy, and input route.
The Automator scores complete recipes using held-out adaptation patients.  A
shared evidence map forms a compact weighted combination, while the reliability check
protects a predefined default recipe when the comparison is uncertain.
Participating recipes are refit on the adaptation
cohort before evaluation begins.}
\label{fig:framework}
\end{figure*}

\section{Method}
\label{sec:method}

\method{} separates two responsibilities that ordinary fine-tuning treats as
one decision.  The \emph{Adapter} defines and executes a finite library of
adaptation recipes.  The \emph{Automator} uses the small adaptation cohort to
weight complete recipes for the current task.  Figure~\ref{fig:framework}
shows this interface.  The separation allows the recipe library to grow
without assuming that a larger search space can be ranked reliably from a few
patients.

\subsection{Adapter}
\label{sec:adapter}

\paragraph{Four candidate spaces.}
The Adapter starts from four finite, explicitly enumerated spaces:
\begin{itemize}[leftmargin=*]
\item $\C_t\subseteq\C$ is the \emph{checkpoint space}.  It contains source
representations compatible with the target signal geometry, including the
general pretrained encoder, task-specialized encoders, and declared fixed
checkpoint combinations.
\item $\mathcal U_t$ is the \emph{update space}.  Its choices are source-head
reuse, a new head on a frozen encoder, partial fine-tuning, LoRA, and full
fine-tuning.  Each choice specifies exactly which parameters may change.
\item $\mathcal D_t$ is the \emph{target-data-policy space}.  A policy fixes
class or patient weighting, the number and type of waveform windows sampled
per patient, and the rule used to aggregate window predictions.
\item $\mathcal M_t$ is the \emph{input-route space}.  Depending on data
availability, it contains waveform only, EHR only, early fusion of waveform
and EHR representations, and late fusion of their predictions.
\end{itemize}
These are spaces of concrete executable choices, not learned embedding spaces.
Unavailable or incompatible choices are removed before adaptation.  The valid
recipe library for task $t$ is therefore
\begin{equation}
  \R_t=\left\{(c,u,d,m)\in
  \C_t\times\mathcal U_t\times\mathcal D_t\times\mathcal M_t:
  \operatorname{valid}_t(c,u,d,m)=1\right\}.
  \label{eq:recipe}
\end{equation}
Here $r=(c,u,d,m)$ is one complete recipe.  The validity function checks, for
example, that a waveform checkpoint receives the required channels and that a
fusion route is proposed only when both modalities are present.  The four
choices are kept together because their effects interact.  A specialist may
work when frozen but overfit after full updating, and the useful waveform
checkpoint may change when EHR context is added.

\paragraph{Executing a recipe.}
For recipe $r$, let $S_d(i)$ be the waveform windows selected for patient $i$,
$w_d(y_i)$ its class weight, and $A_d$ its patient-level aggregation rule.
The route $m$ determines whether the predictor consumes $x_{ij}$, $z_i$, or
both.  For an EHR-only route, waveform windows are ignored and $A_d$ is the
identity.  The Adapter fits only the parameter set $\Omega_u(c,m)$ permitted
by update choice $u$:
\begin{equation}
  \hat\omega_t^r=\arg\min_{\omega\in\Omega_u(c,m)}
  \frac{1}{|\A_t|}\sum_{i\in\A_t} w_d(y_i)\,
  \ell\!\left(y_i,
  A_d\!\left(\{f_{c,m}(x_{ij},z_i;\omega):j\in S_d(i)\}\right)\right).
  \label{eq:adapterobjective}
\end{equation}
Here $f_{c,m}$ is the predictor assembled from checkpoint $c$ and input route
$m$, and $\ell$ is the supervised target loss (binary cross-entropy for the
binary tasks studied here).  The loss is averaged over patients rather than
windows, so a long recording does not contribute more independent evidence.
Equation~\ref{eq:adapterobjective} produces one fitted patient-risk predictor
$\hat f_t^r$ for every valid recipe.

\subsection{Automator}
\label{sec:automator}

The Automator runs separately for every target task and returns weights over
$\R_t$.  It uses three inputs: a task and shift prior, internal evidence from
the adaptation cohort, and an estimate of how reliable that evidence is.  The
same Automator parameters are shared across tasks; only its inputs change.

\paragraph{Task and shift prior.}
For each valid recipe $r=(c,u,d,m)$, the prior combines the clinical relation
between source checkpoint $c$ and target task $t$ with label-free compatibility
between the source and target representations produced by route $m$:
\begin{equation}
  \pi_t(r)=
  \alpha\,\operatorname{sim}_{\mathrm{clinical}}(t,c)
  +(1-\alpha)\left[1-\Delta_{\mathrm{input}}(t,c,m)\right].
  \label{eq:shiftprior}
\end{equation}
Here $\alpha\in[0,1]$ is a global mixing coefficient and
$\operatorname{sim}_{\mathrm{clinical}}(t,c)\in[0,1]$ is the Jaccard overlap
between fixed source- and target-task concept sets.  Thus it is one when the
two sets match and zero when they share no concept.  The quantity
$\Delta_{\mathrm{input}}(t,c,m)\in[0,1]$ measures how easily unlabeled source
patients can be distinguished from target adaptation patients under checkpoint
$c$ and route $m$; zero means indistinguishable and one means completely
separated.  Fixed checkpoint combinations average their component priors.
Neither term uses clinical outcome labels.

\paragraph{Evidence from adaptation patients.}
No additional patient cohort is reserved for this step.  Instead, $\A_t$ is
partitioned into patient-level internal folds.  For each fold, recipe $r$ is
fitted on the other folds and predicts the patients in the held-out fold.  Let
$\hat p'_{i,r}$ denote the resulting prediction for patient $i$; the prime
indicates that patient $i$ was excluded from the corresponding fit.  Repeating
this process yields exactly one internally held-out prediction for every
patient in $\A_t$, without using $\E_t$.  Let $\mathcal S_t$ denote the
task-appropriate selection score, oriented so that larger is better.  The
adaptation utility is
\begin{equation}
  Q_t(r)=
  \mathcal S_t\!\left(\{(y_i,\hat p'_{i,r}):i\in\A_t\}\right).
  \label{eq:utility}
\end{equation}
The score $\mathcal S_t$ may be any prespecified validation utility appropriate
to the task.  Crucially, $Q_t$ scores the complete tuple $(c,u,d,m)$; the four
dimensions are not ranked independently.

\paragraph{Combining recipes from adaptation evidence.}
A high $Q_t(r)$ from a few patients does not by itself justify replacing the
predefined default recipe.  The recipe feature vector $\phi_t(r)$ contains its
internally held-out utility, the fraction of leave-one-positive-patient
recomputations that preserve its advantage, its prior $\pi_t(r)$, its average
prediction disagreement with other recipes, its number of updated parameters,
and one-hot indicators of $(c,u,d,m)$.  The task evidence vector $h_t$ contains
the numbers of positive and negative adaptation patients, adaptation
prevalence, and the fraction of recipes with identifiable internal scores.  A
linear scoring map with coefficients $\theta$ produces
\begin{equation}
  a_t(r)=\theta^\top
  \left[\phi_t(r);\,\phi_t(r)\otimes h_t\right],
  \qquad
  \widetilde{w}_t=\mathcal{N}_{\lambda}\!\left((a_t(r))_{r\in\R_t}\right),
  \label{eq:selection}
\end{equation}
where $\mathcal N_{\lambda}$ is the globally fixed regularized map from recipe
scores to recipe-combination weights.  The coefficients $\theta$, the choice
between nonnegative weights and a regularized signed correction, and the regularization
$\lambda$ are system-level structural choices.  Their selection scope is
reported in Section~\ref{sec:experiments}; once a configuration is chosen,
these values remain fixed while its numerical task-specific weights are
estimated from $\A_t$.  A one-hot vector is the special case of selecting one
recipe.

Finally, a fixed reliability rule maps positive-patient support, the bootstrap
probability that the weighted combination improves over the default, and the
leave-one-positive-patient stability defined above to $\rho_t\in[0,1]$.  The
reference recipe $r_{0,t}$ is a simple valid recipe
declared as the default before target evaluation.  Let $e_{r_{0,t}}$ be its
one-hot weight vector.  The final weights and prediction
are
\begin{equation}
  w_t=\rho_t\widetilde{w}_t+(1-\rho_t)e_{r_{0,t}},
  \qquad
  \hat f_t(X_i,z_i)=\sum_{r\in\R_t}w_t(r)\hat f_t^r(X_i,z_i).
  \label{eq:recipecombination}
\end{equation}
If the adaptation comparison is not identifiable, $\rho_t=0$ and the default
recipe is preserved.  Otherwise, the Adapter refits every recipe with nonzero
weight on all of $\A_t$.  The weights, fitted recipes, and patient aggregation
rules are then locked before $\E_t$ is scored.  Adding a checkpoint, update
rule, data policy, or modality route expands $\R_t$ without changing the
Automator interface.
\section{Experimental Setup}
\label{sec:experiments}

\subsection{Datasets And Baselines}

The evaluation separates hospital transfer from more distant changes in care
setting and acquisition.  MIMIC-III supplies the source waveform checkpoints
~\citep{johnson2016mimic3,goldberger2000physiobank}.  ALOTT is the primary
method-development target because it changes the hospital while preserving the
ICU setting.  It tests whether AutoAdapt can reuse source physiology for new
outcomes without conflating the result with a completely different acquisition
protocol~\citep{lawrence2025alott}.  MIMIC-IV is a second development benchmark and provides the sharper
scarcity test through its much smaller  EHR and waveform cohort
~\citep{johnson2023mimic4,moody2022mimic4waveform}.
The remaining datasets test where the same decision rule remains useful.
MC-MED changes the setting from intensive care to emergency care
~\citep{kansal2025mcmed}.  CPSC 2018
and Ningbo replace continuous bedside monitoring with diagnostic ECGs acquired
under different protocols~\citep{liu2018cpsc,zheng2020chapman}.  We report
results by dataset because these outcomes and evidence regimes are not
interchangeable.  Every included task has a reproducible prediction time and a
patient-level outcome.  
We consider the following baselines: (1) standard few-shot baselines, including a
logistic, prototype, or 5-NN head on a frozen encoder, partial fine-tuning,
LoRA, or full fine-tuning~\citep{hu2022lora},  (2) existing few-shot adaptation baselines: PRAM, PRAM-MI, and OTTEHR~\citep{jeong2026pram,li2026ottehr}, 
and (3) recent automatic composition rules.  LoraHub learns convex candidate weights from 
adaptation predictions~\citep{huang2024lorahub}; AdaMerging estimates weights
by entropy minimization on unlabeled target predictions~\citep{yang2024adamerging};
and HOSO selects the blend between the source route and a target-updated route
using held-out adaptation evidence~\citep{vorster2026hoso}.

\begin{table*}[!t]
\centering
\footnotesize
\setlength{\tabcolsep}{0.8pt}
\caption{MIMIC-III$\rightarrow$ALOTT few-shot comparison on six clinically
central tasks selected for clinical relevance, patient support, waveform
coverage, and task difficulty.  Values are patient-level AUROC.  Every method
uses the same 32 positive and 32 negative adaptation patients per task.}
      \vspace{.5em}
\label{tab:external-raw-waveform}
\begin{tabular}{L{4.7cm}*{7}{C{1.25cm}}}
\toprule
Method & Ventr. & \shortstack{Crit.\\rhythm} & \shortstack{Oxy.\\crisis} & Short VT & \shortstack{Pause/\\asystole} & ST shift & Mean \\
\midrule
\multicolumn{8}{l}{\textit{Fixed recipe baselines}}\\[-2pt]
Frozen source encoder + target head & .512 & .556 & .515 & .516 & .567 & .622 & .548 \\
Full fine-tuning & .642 & .663 & .560 & .677 & .654 & .731 & .655 \\
LoRA & .640 & .662 & .557 & .672 & .656 & .716 & .651 \\
\midrule
\multicolumn{8}{l}{\textit{Existing adaptation methods}}\\[-2pt]
PRAM \citep{jeong2026pram} & .517 & .515 & .516 & .545 & .487 & .518 & .516 \\
PRAM-MI \citep{jeong2026pram} & .527 & .514 & .524 & .545 & .487 & .526 & .520 \\
OTTEHR \citep{li2026ottehr} & .552 & .478 & .629 & .530 & .581 & .547 & .553 \\
LoraHub \citep{huang2024lorahub} & .634 & .661 & .514 & .700 & .657 & .770 & .656 \\
AdaMerging \citep{yang2024adamerging} & .549 & .627 & .514 & .600 & .642 & .716 & .608 \\
HOSO \citep{vorster2026hoso} & .636 & .662 & .515 & .668 & .654 & .753 & .648 \\
\midrule
\textbf{\method{}} & \textbf{.677} & \textbf{.684} & \textbf{.612} & \textbf{.729} & \textbf{.692} & \textbf{.810} & \textbf{.701} \\
\bottomrule
\end{tabular}
\vspace{-1em}
\end{table*}

\subsection{Adaptation Protocol}

We convert each waveform into a five-minute, 125-Hz input on the masked channel
grid used by the source model. EHR features use only measurements available before the prediction time.  Patients are disjoint across splits whenever patient
identifiers are available. 
For the binary experiments, the generic score $\mathcal S_t$ is instantiated
as patient-level AUROC.  We instantiate $\Delta_{\mathrm{input}}$ as
$2|A_{\mathrm{domain}}-1/2|$, where $A_{\mathrm{domain}}$ is the
cross-validated discrimination score of a balanced linear classifier that
separates source from adaptation-patient embeddings.  The clinical concept
sets used in the Jaccard term are fixed before task evaluation.

Each ALOTT task uses the same 64 adaptation patients for every method, with 32
positive and 32 negative patients.  Training is capped at eight waveform
windows per patient so that recording duration cannot substitute for patient
support.  Recipe heads and task-specific evidence are learned only from these
adaptation patients. The EHR baseline follows the same label
budget.  For the reported ALOTT analysis, combination weights and
calibration are fitted on adaptation patients.

The available input-route space depends on the dataset.  ALOTT exposes EHR-only,
waveform-only, early-fusion, and late-fusion recipes; the selected combinations
contain a fusion route in four of the six tasks in Table~\ref{tab:external-raw-waveform}.
MC-MED likewise allows EHR, waveform, and fusion, with EHR as the declared
default recipe.  The primary MIMIC-IV comparison fixes the route to waveform-only,
while its EHR model is reported as a separate target-site baseline.  CPSC and
Ningbo contain no structured EHR route, so their recipe libraries are
waveform-only.  Thus $\mathcal M_t$ is selected where multiple modalities are
available and becomes a singleton where they are not.

\begin{table*}[!t]
\centering
\footnotesize
\setlength{\tabcolsep}{5.8pt}
\caption{MIMIC-IV scarce-label stress test using patient-level AUROC averaged
over five evaluation runs. Every method uses the same adaptation and
evaluation patients in each run. Underlining marks the strongest non-AutoAdapt
result and boldface marks the highest result.}
      \vspace{.5em}
\label{tab:maincomparison}
\begin{tabular}{L{4.6cm}rrrrrrr}
\toprule
Method & Acute & Circ. & SOFA & Delir. & AKI & Mort. & Mean \\
\midrule
\multicolumn{8}{l}{\textit{Fixed recipe baselines}}\\[-2pt]
Frozen source encoder + target head & .545 & .596 & .640 & .547 & .434 & .637 & .567 \\
Full fine-tuning & .538 & .585 & .591 & .519 & .465 & .665 & .560 \\
LoRA & .564 & .586 & .606 & .513 & .477 & \underline{.685} & .572 \\
\midrule
\multicolumn{8}{l}{\textit{Existing-work baselines}}\\[-2pt]
PRAM \citep{jeong2026pram} & \underline{.574} & .605 & \underline{.669} & \underline{.548} & .468 & .649 & \underline{.585} \\
PRAM-MI \citep{jeong2026pram} & .565 & .601 & .657 & .544 & .464 & .650 & .580 \\
OTTEHR \citep{li2026ottehr} & .486 & .465 & .583 & .495 & \underline{.508} & .576 & .519 \\
LoraHub \citep{huang2024lorahub} & .542 & .649 & .650 & .540 & .405 & .582 & .561 \\
AdaMerging \citep{yang2024adamerging} & .553 & .614 & .586 & .539 & .414 & .613 & .553 \\
HOSO \citep{vorster2026hoso} & .534 & \underline{.660} & .659 & .536 & .415 & .582 & .564 \\
\midrule
\multicolumn{8}{l}{\textit{Ours}}\\[-2pt]
\textbf{\method{}} & \textbf{.847} & \textbf{.934} & \textbf{.887} & \textbf{.842} & \textbf{.904} & \textbf{.847} & \textbf{.877} \\
\bottomrule
\end{tabular}
  \vspace{-1em}
\end{table*}
\section{Experimental Results}
\label{sec:results}

\subsection{Main Results on ALOTT ICU Transfer}
\label{sec:alott-primary}

The ALOTT registry supplies twenty duration-qualified monitor-event tasks at a
new ICU.  Table~\ref{tab:external-raw-waveform} presents six clinically central
tasks, and Appendix Table~\ref{tab:alott-remaining} reports the remaining
fourteen.
Every method receives the same 64 adaptation patients for each task.  The
comparison therefore measures how the same small cohort is used.

The strongest displayed baseline is LoraHub at .656 mean AUROC.  In contrast,
\method{} uses adaptation evidence to construct a task-specific weighted
combination.

\method{} uses one shared evidence map across the six reported tasks and
achieves a mean AUROC of .701.  It leads the displayed baselines on five of six
tasks, while OTTEHR is stronger on oxygenation crisis.  Across the full
twenty-task roster, \method{} reaches .686 mean AUROC
versus .652 for the previous same-cohort reference.  The consistent
gains support the main mechanism: adaptation evidence is more useful when it
controls a regularized combination than when it makes a winner-take-all choice.
ALOTT is used for method development; the later datasets test the same
selection principle under larger acquisition changes.

\subsection{Extreme Few-Shot Selection On MIMIC-IV}
\label{sec:mainresults}

MIMIC-IV tests the selector where adaptation evidence is most fragile: across
five fixed patient partitions and six tasks, each adaptation split contains
only one to six positive patients, and every recipe is selected using only
those adaptation patients.  Table~\ref{tab:maincomparison} shows that
\method{} leads the displayed baselines on every task.  Its shared evidence
map combines patient-held-out adaptation performance with task similarity,
distribution shift, target prediction geometry, and positive-patient support,
without learning task- or fold-specific coefficients.  This structure keeps
weak evidence from dominating merely because many correlated waveform windows
are available.  The main finding is that reliable task-specific combinations
can still be selected when independent positive patients, rather than raw
window count, are the limiting resource.

\begin{figure}[!t]
  \centering
  \begin{minipage}[t]{0.32\linewidth}\centering
    \includegraphics[page=14,width=\linewidth]{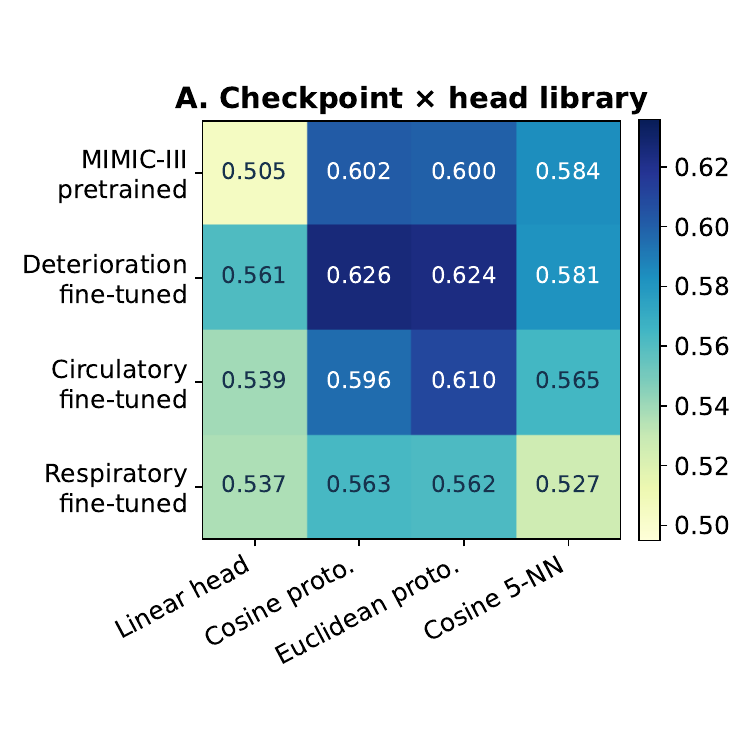}
  \end{minipage}\hfill
  \begin{minipage}[t]{0.32\linewidth}\centering
    \includegraphics[page=15,width=\linewidth]{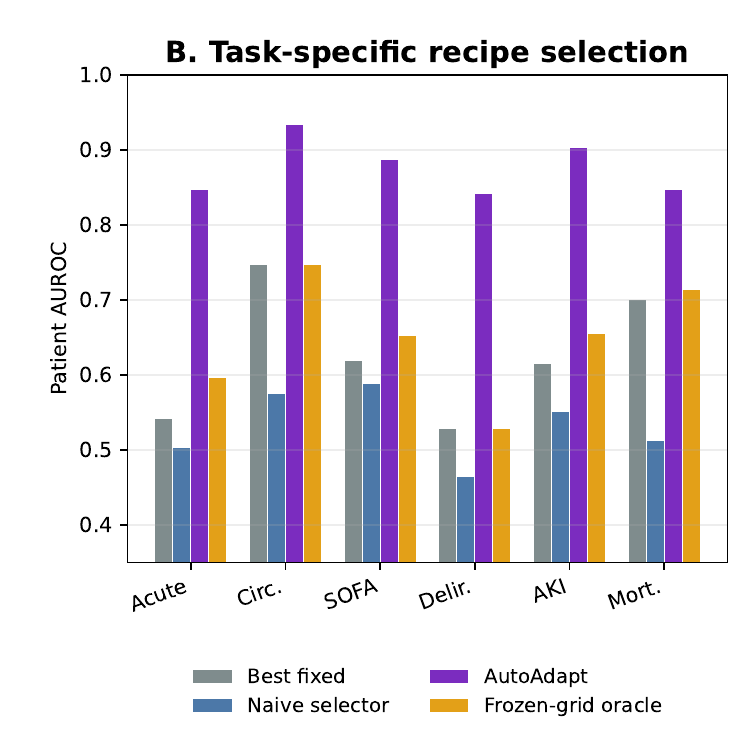}
  \end{minipage}\hfill
  \begin{minipage}[t]{0.32\linewidth}\centering
    \includegraphics[page=16,width=\linewidth]{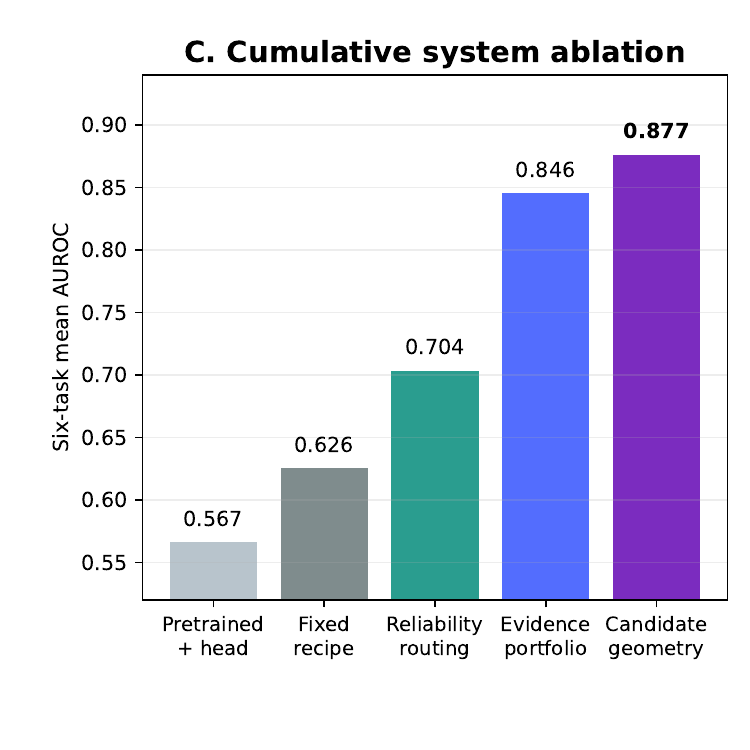}
  \end{minipage}
  \vspace{-1em}
  \caption{Checkpoint and recipe analysis on MIMIC-IV.  Panel A tests source
  checkpoint and head pairings.  Panel B compares a fixed recipe, naive
  adaptation-only selection, and the reliability-aware Automator.  Panel C
  adds the Adapter and Automator components cumulatively.}
  \label{fig:selected-checkpoint-head}
\end{figure}

\subsection{Cross-Dataset Transfer On MC-MED And Diagnostic ECG}
\label{sec:crossdataset}

\begin{figure*}[!t]
  \centering
  \includegraphics[width=0.96\linewidth]{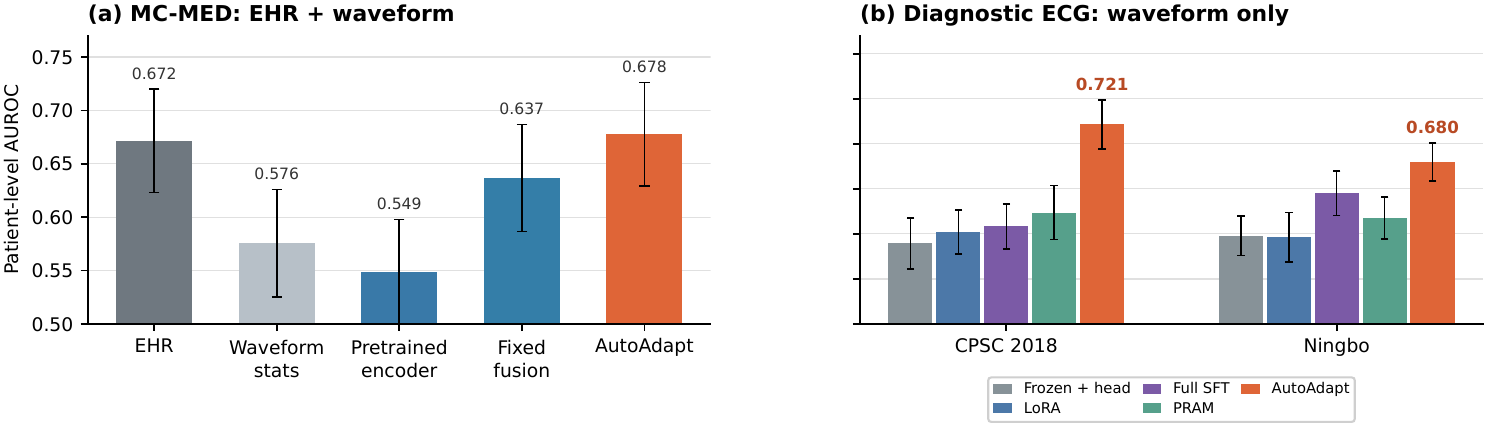}
    \vspace{-1em}
  \caption{Cross-dataset comparisons.  Panel A reports MC-MED macro AUROC over
  four tasks with sufficient patient support, comparing local EHR, waveform
  transfer, fixed fusion, and \method{}.  Panel B reports macro AUROC over five
  shared diagnostic ECG tasks in CPSC and Ningbo.  Methods use the same target
  folds within each dataset.  The reference line marks chance AUROC.}

  \label{fig:crossdataset-baselines}
\end{figure*}

The recipe library is useful only if it avoids forcing source physiology onto
an incompatible target.  MC-MED tests transfer from ICU monitoring to emergency
care, while CPSC and Ningbo replace continuous bedside signals with diagnostic
ECGs.  As Figure~\ref{fig:crossdataset-baselines} shows, \method{} retains the
local EHR route on MC-MED unless adaptation evidence supports waveform fusion,
but makes greater use of source waveform features when ECG structure remains
compatible on CPSC and Ningbo.  The cross-dataset finding is therefore not
that source representations should always be reused, but that adaptation should
depend on whether their signal structure survives the target shift.  CPSC does
not provide patient identifiers, so patient-disjoint splitting cannot be
guaranteed for that dataset.

\subsection{Checkpoint And Prediction-Head Analysis}
\label{sec:focused-analysis}

A simpler system could choose a checkpoint and prediction head independently.
Figure~\ref{fig:selected-checkpoint-head}A tests this shortcut by pairing four source encoders with
four frozen heads on the same target
patients.
Mean AUROC ranges from .505 to .626 across these pairs.  The deterioration
checkpoint with a cosine prototype is strongest on average, but the preferred
pair changes across tasks.  More importantly, the relative order of heads
changes with the representation.  This interaction rules out independent
component ranking as an explanation for \method{} and motivates selection over
complete recipes.

\begin{figure}[!t]
  \centering
  \includegraphics[width=0.98\linewidth]{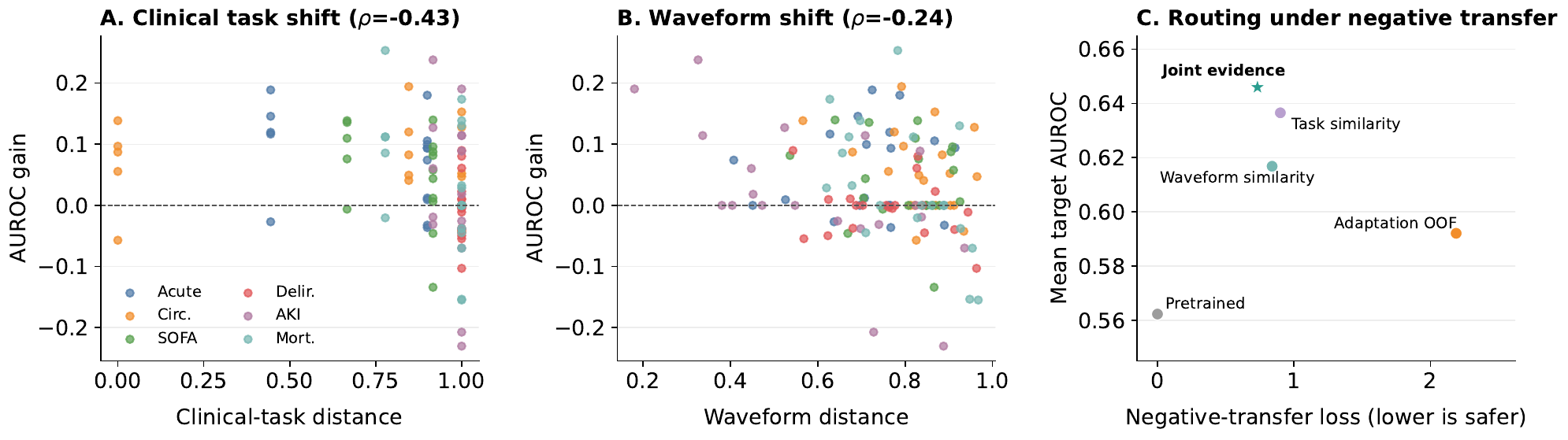}
  \caption{Checkpoint routing under distribution shift.  Panels A and B show
  how transfer gain changes with task distance and waveform distance.  Panel C
  compares routing rules.  Every point uses the same target patients.
  Evaluation labels measure transfer gain but do not select a checkpoint.}
  \label{fig:joint-shift-analysis}

\end{figure}

\begin{figure}[!t]
  \centering
  \includegraphics[width=.85\linewidth]{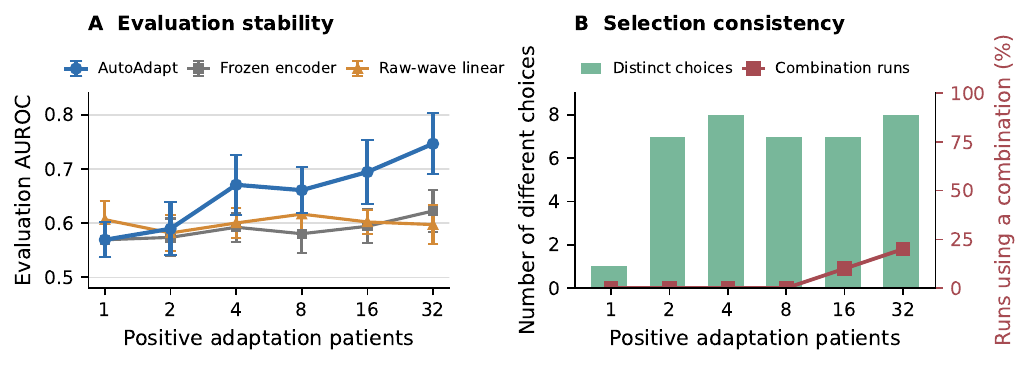}

  \caption{Stability on ALOTT new-pressor prediction as the adaptation panel
  grows from one to 32 positive patients, with an equal number of negative
  patients.  Panel A reports evaluation AUROC over ten repeated panels against
  two fixed baselines.  Panel B reports the number of distinct selections and
  the fraction using a recipe combination.  } 
  \label{fig:positive-patient-budget}

\end{figure}

\subsection{Reliability-Aware Selection Analysis}

Selecting the largest internal score is the most direct alternative to the
Automator.  We compare that rule with one strong fixed recipe and the complete
reliability-aware selector while holding the candidate library and target
patients constant in Figure~\ref{fig:selected-checkpoint-head}B.
Naive selection often overreacts to an unstable few-shot ranking and can be
worse than keeping a fixed recipe.  Task and shift priors narrow the plausible
choices, while the reliability check rejects changes that lack patient-level
support.  Weighted combinations and target prediction geometry then preserve
complementary candidates without trusting a single noisy winner.  The central
finding is therefore not that a larger search space is inherently better, but
that reliability-aware evidence makes a diverse search space usable.

\subsection{Adapter and Automator Ablation Study}

Figure~\ref{fig:selected-checkpoint-head}C presents the cumulative ablation and tests
whether a larger library alone explains the result.  The stages start with the
general pretrained encoder, add the strongest fixed recipe, introduce task and
shift routing, and finally apply complete recipe selection with the reliability
gate.
Each component produces a cumulative improvement, but expanding the Adapter
alone explains only the first gain.  The larger improvement appears when the
Automator routes among those alternatives using reliability and shift evidence.
Keeping checkpoint, update-method, and data-policy evidence separate prevents
one strong signal from being diluted, while candidate-level target geometry
distinguishes redundant predictions from complementary ones.  The ablation
therefore supports the intended division of labor: the Adapter supplies useful
alternatives, and the Automator decides when and how to use them.

\subsection{Clinical-Task And Waveform-Shift Analysis}

The prior assumes that clinical relation and signal compatibility describe
different parts of transfer.  We test this assumption with a fixed target head
and data policy so that Figure~\ref{fig:joint-shift-analysis} isolates source
checkpoint routing.
Clinical task distance correlates at $-.43$ with specialist transfer gain,
while waveform distance correlates at $-.24$.  Each association is informative
but incomplete.  Combining both signals with adaptation evidence raises mean
AUROC from .562 to .646 and reduces harmful transfer by 67\% relative to
adaptation-only routing.  The task relation identifies source knowledge that
could matter, while the target waveform reveals whether that knowledge remains
compatible with the new acquisition distribution.

\subsection{Checkpoint-Pool Composition Analysis}

We grow the source pool from the general pretrained encoder to test whether
checkpoint count alone helps.  The deterioration specialist raises mean
routing AUROC from .562 to .641, and the circulatory specialist raises it to
.646.  The respiratory specialist is never selected and adds no gain.  The
library therefore benefits from complementary physiological knowledge, not
from accumulating redundant checkpoints.

\subsection{Adaptation-Patient Budget And Stability Analysis}

To test the MIMIC-IV failure mode directly, we vary positive-patient support
rather than total window count.  Ten balanced ALOTT adaptation panels are drawn
at each budget from one to 32 positives while evaluation remains fixed
(Figure~\ref{fig:positive-patient-budget}).  With one positive patient, the
support rule always retains the pretrained default.  From four positives
onward, \method{} exceeds both fixed baselines and reaches .747 mean AUROC at
32.  Different panels still select different recipes, showing that more
independent outcomes strengthen the gate without making the choice
deterministic.

\section{Conclusion}

Adapting a clinical model with a few labeled patients is a decision problem as
much as an optimization problem.  The useful source checkpoint, update rule,
data policy, and modality route change across tasks, while the cohort used to
compare them may contain only a handful of positive outcomes.  \method{}
addresses these two constraints with an extensible Adapter and a
patient-reliability-aware Automator.  The experiments show that this separation
turns heterogeneous source models into useful task-specific predictors, while
the predefined default limits unsupported transfer under larger shifts.  Reliable recipe
selection provides a practical path for reusing clinical models where target
labels are scarce and no single adaptation rule is consistently appropriate.


\bibliographystyle{style/icml2025}
\bibliography{999_reference}

\clearpage
\appendix
\section{Detailed Task And Data Contracts}
\label{app:data}

\begin{table*}[t]
\centering
\scriptsize
\setlength{\tabcolsep}{4pt}
\caption{The fourteen ALOTT tasks not included in the six-task main panel.
Values are patient-level AUROC on the same evaluation cohorts used for the
main comparison.}
\label{tab:alott-remaining}
\begin{tabular}{L{6.0cm}rrr}
\toprule
Task & Prior reference & \method{} & Gain \\
\midrule
AF vs sustained VT/VF alarm proxy & .727 & .793 & +.066 \\
Critical rhythm vs short VT severity proxy & .612 & .655 & +.043 \\
Critical rhythm vs ST-shift severity proxy & .663 & .697 & +.033 \\
Critical rhythm vs VPC severity proxy & .597 & .647 & +.050 \\
Critical vs noncritical electrical proxy & .569 & .617 & +.047 \\
Pause vs sustained VT/VF proxy & .595 & .653 & +.057 \\
Short VT vs nonrun ectopy proxy & .593 & .629 & +.035 \\
Short VT vs pause proxy & .639 & .660 & +.021 \\
VPC vs pause proxy & .649 & .704 & +.056 \\
VPC vs short VT proxy & .645 & .659 & +.014 \\
VPC vs ST-shift proxy & .682 & .733 & +.051 \\
VT family vs pause proxy & .641 & .689 & +.048 \\
Future VPC-burden escalation proxy & .659 & .674 & +.015 \\
Respiratory crisis proxy & .560 & .612 & +.052 \\
\bottomrule
\end{tabular}
\end{table*}

\begin{table}[h]
\centering
\footnotesize
\setlength{\tabcolsep}{2.5pt}
\caption{Independent patient support in the five MIMIC-IV runs.  Ranges show
variation across patient partitions.  Evaluation counts include patients with
a usable prediction.  Waveform row counts are omitted because repeated
measurements are not independent events.}
\label{tab:cohort}
\begin{tabular}{lcrrrrrr}
\toprule
 & & \multicolumn{2}{c}{All eligible} & \multicolumn{2}{c}{Adaptation/run} & \multicolumn{2}{c}{Evaluation/run} \\
\cmidrule(lr){3-4}\cmidrule(lr){5-6}\cmidrule(lr){7-8}
Task & O/G/H (h) & Patients & Positive & Patients & Positive & Patients & Positive \\
\midrule
Acute & 6/0/6 & 87 & 30 & 17--18 & 6 & 69--70 & 24 \\
Circ. & 6/0/6 & 139 & 26 & 27--29 & 5--6 & 110--112 & 20--21 \\
SOFA & 6/6/24 & 147--148 & 25--26 & 29--30 & 5 & 117--119 & 20--21 \\
Delir. & 6/3/9 & 91 & 22 & 17--19 & 4--5 & 72--74 & 17--18 \\
AKI & 6/1/23 & 46--47 & 19 & 9--10 & 4 & 36--37 & 15 \\
Mort. & 6/6/24 & 157--158 & 8 & 31--32 & 1--2 & 125--127 & 6--7 \\
\bottomrule
\end{tabular}
\end{table}

Task labels are constructed from clinical events after the waveform cutoff.
In Table~\ref{tab:cohort}, O/G/H gives the observation period, blank period,
and prediction horizon in hours.  Acute deterioration combines new organ
support with death.  Circulatory deterioration uses new pressor initiation.
SOFA requires a future component increase of at least two
\citep{vincent1996sofa}.  Delirium requires a new positive CAM record
\citep{ely2001camicu}.  AKI uses incident urine output KDIGO stage 2 or 3
\citep{khwaja2012kdigo}.  Dynamic mortality uses future in hospital death.
Windows are masked when the future state cannot be observed.

The primary MIMIC-IV input is waveform only.  Every interval is resampled to
125\,Hz and mapped to the fixed 62 channel vocabulary.  Missing channels are
zero with an explicit presence mask.  We corrected inconsistent aliases for
respiration and aVR before running the reported comparison. 
Labels and patient partitions remain unchanged.

\section{Reproducibility Details}
\label{app:repro}

\begin{table}[t]
\centering
\footnotesize
\setlength{\tabcolsep}{3.6pt}
\caption{Source checkpoint library.  Specialized checkpoints start from the
general pretrained model and receive supervised MIMIC-III training.  A source
classifier is reused only when its label matches the target contract.}
\label{tab:sourcelibrary}
\begin{tabular}{L{2.2cm}L{4.3cm}rL{3.4cm}}
\toprule
Checkpoint & MIMIC-III training state & Saved model & Target use \\
\midrule
Pretrained & Masked time and frequency pretraining at selected step 1,300 & 37,699,612 & Encoder with target fitted head \\
Deterioration & Pretrained start with escalation or death supervision at epoch 8 & 37,700,125 & Encoder with aligned acute source head \\
Circulatory & Pretrained start with five hemodynamic labels at epoch 5 & 37,702,177 & Encoder with aligned pressor source head \\
Respiratory & Pretrained start with five respiratory labels at epoch 15 & 37,702,177 & Encoder only for main targets \\
\bottomrule
\end{tabular}
\end{table}

\begin{table}[t]
\centering
\scriptsize
\setlength{\tabcolsep}{3.0pt}
\caption{Target datasets and evaluation scope.  Support gives positive
evaluation patients after waveform quality control for ALOTT and MC-MED.  The
MIMIC-IV entry gives the full eligible range.}
\label{tab:datasetroles}
\begin{tabular}{L{1.55cm}L{2.15cm}L{2.35cm}L{3.7cm}}
\toprule
Target & Setting and input & Positive support & Purpose \\
\midrule
ALOTT & ICU with EHR and bedside waveforms & 16--779 & Primary  64-patient transfer \\
MIMIC-IV & ICU waveform with EHR extension & 8--30 & Scarce label stress test \\
MC-MED & ED with EHR and bedside waveforms & 23--162 & Reliability of the default recipe \\
CPSC and Ningbo & Diagnostic 12 lead ECG & Native folds & Acquisition shift \\
\bottomrule
\end{tabular}
\end{table}

\begin{table}[t]
\centering
\footnotesize
\setlength{\tabcolsep}{3.5pt}
\caption{Target adaptation performed by each comparison.  Candidate models and
combination weights use labels from the declared adaptation patients.}
\label{tab:baselinecontract}
\begin{tabular}{L{2.6cm}cL{3.1cm}L{4.2cm}}
\toprule
Method & Target labels & Encoder update & Target operation \\
\midrule
\multicolumn{4}{l}{\textit{Traditional baselines}}\\[-2pt]
Source head & No & None & Reuse the original source classifier unchanged \\
Frozen + head & Yes & None & Fit a new logistic head on frozen embeddings \\
Frozen + cosine prototype & Yes & None & Compare with adaptation class centroids using cosine similarity \\
Frozen + Euclidean prototype & Yes & None & Compare with adaptation class centroids using Euclidean distance \\
Frozen + cosine 5-NN & Yes & None & Vote among five nearby adaptation examples \\
Partial fine tuning & Yes & Last two blocks & Fit the final blocks and a new head \\
LoRA & Yes & Rank 8 adapters & Fit low rank attention updates and a new head \\
Full fine tuning & Yes & Active path & Update the classification path and new head \\
\midrule
\multicolumn{4}{l}{\textit{Existing work baselines}}\\[-2pt]
PRAM & Yes & None & Mix predictions retrieved from adaptation patients \\
PRAM-MI & Yes & None & Use mutual-information-weighted Euclidean retrieval \\
OTTEHR & Yes & None & Transport frozen validation features toward the adaptation reference \\
LoraHub & Yes & None & Learn convex candidate weights from adaptation predictions \\
AdaMerging & No & None & Minimize prediction entropy on unlabeled target examples \\
HOSO & Yes & None & Select a source and target-updated prediction blend \\
\midrule
\multicolumn{4}{l}{\textit{Ours}}\\[-2pt]
\method{} & Yes & Recipe dependent & Weight complete checkpoint, update, data-policy, and input-route recipes \\
\bottomrule
\end{tabular}
\end{table}

\paragraph{Evaluation contract.}
Patient level AUROC is the primary metric.  The MIMIC-IV table uses five
executions recorded in a fixed manifest.  Each execution assigns about 20\%
of patients to adaptation.  The remaining patients are used for evaluation.
Every MIMIC-IV analysis reuses these assignments and seeds.  CPSC uses five
native folds.  Duplicate
records remain in one fold, but patient disjointness cannot be established
because identifiers are unavailable.  Ningbo uses three participant disjoint
folds with one ECG per participant.  Their results are reported separately.
The primary MIMIC-IV comparison uses waveform input.  Structured EHR defines
the outcome but does not enter the model.  The multimodal analysis adds only
features observed before the prediction cutoff.

\paragraph{Model and optimization.}
The source input is the final five minutes of the observation window on 62
canonical physical channels at 125\,Hz, giving
$x\in\mathbb{R}^{62\times37{,}500}$.  Missing channels are zero-filled with a
presence mask.  A magnitude FFT companion uses 250 samples and expands the
input to 124 internal channels.  Raw physical units are retained in the time
branch.  Only FFT segments are standardized.  Concatenating 25 samples from
each internal channel yields 1,500 tokens.  TF-dual contains eight Transformer
blocks with width 512.  It uses eight attention heads, rotary attention, and an
MLP width of 2,048.  Mean pooling produces a 512 dimensional embedding.  The
checkpoint contains 37.7M parameters.  The active classification path uses
26.81M of them.

The source manifest contains 398,204 one hour MIMIC-III windows.  The loader
exposes 4,778,448 five minute segments without overlap.  Pretraining uses 3,500
optimizer steps with global batch 512.  The deterioration, respiratory, and
circulatory source tasks use 484,667, 551,232, and 578,274 windows.  Their
patient counts are 5,693, 4,421, and 4,955 after validation holdout.  Partial
fine tuning updates the final two blocks, the final LayerNorm, and the target
head.  This exposes 6,306,305 parameters.  Rank 8 LoRA updates the attention
projections and target head, which exposes 197,121 parameters.  Full tuning
updates the 26.81M parameter classification path.  Optimizer and schedule
settings belong to the selected recipe.

All runs use the same four TF-dual source states.  The library contains the
general pretrained encoder and three specialized encoders.  Those specialists
were trained for deterioration, circulatory decline, and respiratory decline.
A machine readable manifest records each object name, size, digest, and source
commit.  Result generation verifies the manifest before loading predictions.

\begin{table}[t]
\centering
\small
\setlength{\tabcolsep}{4pt}
\caption{Fixed Automator thresholds.  Minimum support controls eligibility.
A challenger below the preferred count also needs a favorable paired
confidence bound.  Complexity resolves task-utility differences within
$\delta_Q$.  A source head is reused only for an aligned label.}
\label{tab:selectorthresholds}
\begin{tabular}{lrrrr}
\toprule
D2 method & Min. pos. & Preferred pos. & Rank & Updated params. \\
\midrule
Aligned source head & 0 & 0 & 0 & 0 \\
Frozen + target head & 4 & 4 & 1 & head only \\
LoRA & 4 & 8 & 2 & 197,121 \\
Partial fine tuning & 4 & 8 & 3 & 6,306,305 \\
Full fine tuning & 4 & 20 & 4 & 26.81M \\
\bottomrule
\end{tabular}
\vspace{2pt}

\parbox{0.94\linewidth}{\footnotesize For the binary tasks in this study,
$\mathcal{S}_t$ is AUROC, $\delta_Q=.01$, and the paired safety margin is
$\epsilon_Q=.02$ AUROC.  A .01 Brier margin provides the secondary calibration
check.  Recipe selection uses 5,000 patient bootstrap draws.  Main shift
analyses use 2,000 draws.}
\end{table}

The task distance prior uses the fixed concept sets in
Table~\ref{tab:conceptsets}.  Evaluation labels do not alter these tokens.

\begin{table}[t]
\centering
\footnotesize
\setlength{\tabcolsep}{4pt}
\caption{Clinical concepts used for task distance.  Tokens are fixed before
analysis.  The general pretrained encoder has no outcome specific concept
prior, so its set is empty.  It remains eligible through distribution and
adaptation evidence.}
\label{tab:conceptsets}
\begin{tabular}{L{3.1cm}L{8.4cm}}
\toprule
Source or target & Concept set \\
\midrule
Pretrained source & $\varnothing$ \\
Deterioration source & death, mortality, organ support, respiratory, circulatory, renal, ventilation, pressor, RRT \\
Circulatory source & circulatory, pressor, hypotension, MAP, lactate, shock \\
Respiratory source & respiratory, ventilation, intubation, FiO$_2$, PEEP, gas exchange \\
\midrule
Acute target & death, organ support, ventilation, pressor, RRT \\
Circulatory target & circulatory, pressor, hypotension, MAP, lactate, shock \\
SOFA target & organ support, respiratory, circulatory, renal, neurologic, coagulation, liver \\
Delirium target & neurologic, delirium \\
AKI target & renal, AKI, creatinine, urine output \\
Mortality target & death, mortality \\
\bottomrule
\end{tabular}
\end{table}

\paragraph{Patient aggregation.}
Every metric is computed after combining a patient's windows into one risk.
Each recipe fixes its aggregation rule before recipe combination.  Candidate rules include the
mean, maximum, upper quantile, and a bounded temporal summary.  Bootstrap
sampling operates on patients.

\paragraph{Compute and access.}
PhysioNet waveforms are read in WFDB form and resampled once into task
manifests.  Frozen methods reuse cached representations.  Partial tuning, LoRA,
and full tuning use GPU execution.  Machine readable artifacts generate both
the reported values and the figures.

\end{document}